\documentclass{article}

\usepackage[dblblindworkshop, final]{neurips_2025}
\workshoptitle{The First Workshop on Generative and Protective AI for Content Creation}

\usepackage{graphicx}

\usepackage[utf8]{inputenc} 
\usepackage[T1]{fontenc}    
\usepackage{hyperref}       
\usepackage{url}            
\usepackage{booktabs}       
\usepackage{amsfonts}       
\usepackage{nicefrac}       
\usepackage{microtype}      
\usepackage{xcolor}         
\usepackage[status=draft]{fixme} 
\usepackage{enumitem}

\usepackage{listings}
\usepackage{graphicx}
\usepackage{caption}
\usepackage{placeins}
\usepackage{booktabs} 
\usepackage{makecell}  
\usepackage{graphicx}  
\usepackage{amsmath}

\lstdefinelanguage{XML}{
  morestring=[b]",
  stringstyle=\color{green!40!black},     
  morecomment=[s]{<?}{?>}, 
  morecomment=[s]{<!--}{-->},
  moredelim=[s][\color{blue!70!black}]{<}{>},   
  moredelim=[s][\color{blue!70!black}]{</}{>},  
  moredelim=[l][\color{blue!70!black}]{/>},
  identifierstyle=\color{brown!70!black},       
  keywordstyle={}
}

\usepackage{colortbl}
\definecolor{cellbg}{RGB}{245,248,252}

\title{CPT: Controllable \& Editable Design Variations with Language Models}

\author{%
  Karthik Suresh \\
  Adobe \\
  \texttt{karsures@adobe.com}
  \And
  Amine Ben Khalifa\thanks{Work done while at Adobe.} \\
  Atta \\
  \texttt{amine.benkhalifa@gmail.com}
  \And
  Li Zhang \\
  Adobe \\
  \texttt{zhangli@adobe.com}
  \And
  Wei-ting Hsu \\
  Adobe \\
  \texttt{whsu@adobe.com}
  \And
  Fangzheng Wu \\
  Adobe \\
  \texttt{fangzhengw@adobe.com}
  \And
  Vinay More \\
  Adobe \\
  \texttt{vmore@adobe.com}
 \And
  Asim Kadav \\
  Adobe \\
  \texttt{akadav@adobe.com}
}

\begin{document}

\maketitle

\begin{abstract}
Designing visually diverse and high-quality designs remains a manual, time-consuming process, limiting scalability and personalization in creative workflows. We present a system for generating editable design variations using a decoder-only language model – the Creative Pre-trained Transformer (CPT) – trained to predict visual style attributes in design templates (Figure \ref{fig:example}). At the core of our approach is a new representation called Creative Markup Language (CML), a compact, machine-learning–friendly format that captures canvas-level structure, page layout, and element-level details (text, images, and vector graphics), including both content and style. We fine-tune CPT on a large corpus of design templates authored by professional designers, enabling it to learn meaningful, context-aware predictions for attributes such as color schemes and font choices. The model produces semantically structured and stylistically coherent outputs, preserving internal consistency across elements. Unlike generative image models, our system yields fully editable design documents rather than pixel-only images, allowing users to iterate, personalize within a design editor. In experiments, our approach generates contextual color and font variations for existing templates and shows promise in adjusting layouts, all while maintaining design principles.
\end{abstract}

\section{Introduction}
Creating visually appealing content that aligns with brand guidelines is essential for creators, marketers, and businesses. However, producing content at scale remains manual and slow. Typically, expert designers are required to create and adapt templates for different contexts or audiences. As demand for high-quality personalized content grows across platforms, this manual workflow becomes a bottleneck that limits both creativity and productivity. Recent advances in generative AI have introduced powerful multimodal models \citep{achiam2023gpt4, openai2024gpt4o, team2023gemini} and tools capable of producing impressive visual artifacts \citep{lin2023autoposter, gao2023textpainter}, yet many of these systems operate in unstructured domains (e.g., image generation) or require extensive prompt engineering without offering editability or control over specific visual attributes. Other tools provide semi-automated recommendations for fonts \citep{nan2018font, jiang2019visual} or context-aware asset search \citep{kovacs2018context}, but these require manual application and often lack contextual awareness or stylistic coherence throughout designs.

In this work, we propose a new perspective on the design variation problem: treating it as a structured sequence prediction task over a semantically rich representation of design documents. We introduce the Creative Pre-trained Transformer (CPT), a decoder-only language model fine-tuned to predict masked style attributes within editable design templates. Central to this approach is the Creative Markup Language (CML), a compact machine learning-friendly representation that encodes the structure, layout, content and styling details at the canvas level for text, images, and vector graphics, inspired by recent work on multimodal document understanding \citep{ye2023mplug, kikuchi2024multimodal}.

By leveraging powerful large language models (LLMs) fine-tuned on designer-authored content, our approach incorporates rich world knowledge about color, typography, and visual design, allowing the generation of context-aware and stylistically sophisticated variations compared to traditional recommendation-based methods. Unlike traditional generation methods that produce static images, our system produces fully editable design documents that can be rendered, edited, and exported within a commercial design editor.

Our contributions are as follows:
\begin{itemize} [leftmargin=*]
\item We propose a controllable and editable design-variation framework (CPT+CML): users choose which attributes to vary (color, font, layout), CPT predicts only those fields under optional brand constraints, and the resulting documents remain fully editable—unlike traditional LLM/diffusion systems that produce uncontrolled, non-editable outputs.
\item We develop a complete engineering pipeline for converting raw production design documents to and from CML, enabling seamless integration with production design tools and editable output.
\item We design a heuristics-based evaluation pipeline complemented by a GPT-powered filter and design scorer, which ensures generated variations meet aesthetic and usability standards.
\item We demonstrate the effectiveness of our approach on generating stylistically coherent color and font variations for templates from a commercial design editor, and show preliminary results for layout variation on simple templates.
\end{itemize}
\section{Background and Related Work}

Early systems for automatic graphic design relied on hard–coded rules or template retrieval, limiting flexibility and scale \citep{odonovan2014,kovacs2018context}.  Recent work treats layout generation as a learning problem.  \textbf{GAN-based} models such as
LayoutGAN refine randomly initialised element boxes using a wire-frame discriminator \citep{li2019layoutgan}.  \textbf{Latent-variable} approaches
(LayoutVAE) capture the distribution of scene layouts via a VAE, while \textbf{autoregressive transformers} (LayoutTransformer, VTN) model element
sequences directly.  Diffusion variants (LayoutDM) further boost diversity and realism through iterative refinement.
\begin{figure}
    \centering
    \includegraphics[width=1\linewidth]{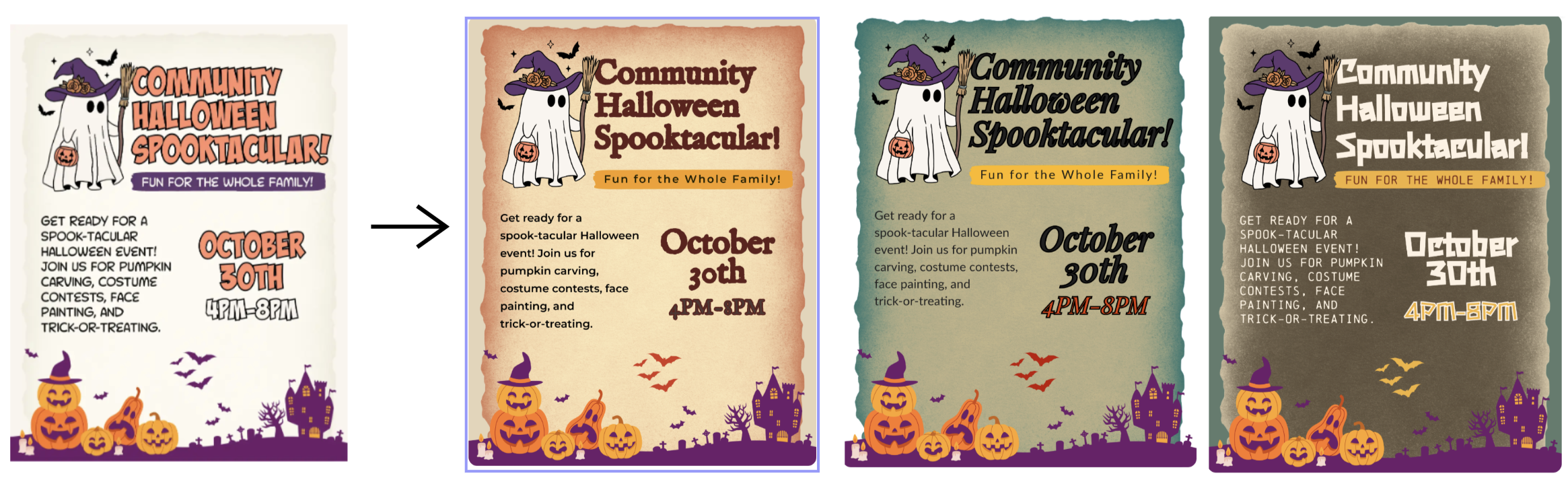}
    \caption{Our CPT model uses the context of the original template (far left) to generate font and color variations.}
    \label{fig:example}
\end{figure}

Most of these methods output abstract bounding boxes and raster previews, leaving detailed styling (color, typography) and editability to the user. To address full-fidelity outputs, CreatiPoster generates multi-layer posters by combining an LLM for structured JSON with a diffusion background model \citep{zhang2025creatiposter}.  Closer to our goal, Shimoda
\textit{et al.} propose a transformer that predicts coherent font-and-color assignments for placeholder text, but it omits images and layout metadata \citep{shimoda2024towards}.  Context-aware recommenders handle specific style subtasks—e.g.\ font pairing \citep{nan2018font,jiang2019visual} or  harmonisation—but treat each element independently and do not scale to full-template variation.

Prior systems differ in output fidelity and editability. LayoutDM and related diffusion/transformer models generate only abstract boxes or raster previews, offering limited control \citep{horita2024retrieval}. AutoPoster produces content-aware posters but as static images without editability \citep{lin2023autoposter}. CreatiPoster improves fidelity with layered JSON plus a diffusion background, yet control remains partial (layers only) \citep{zhang2025creatiposter}. In contrast, CPT outputs fully structured, editable CML documents with fine-grained control (color, font, layout, brand), unifying editability and controllability in a single framework.


Our work unifies these threads: we cast template variation as a \emph{masked sequence prediction} task over a compact, editable markup (CML) and fine-tune a 7B decoder-only LLM (CPT) to jointly predict color, font and layout attributes.  Unlike prior image-centric generators, CPT outputs a fully structured document that designers can open and tweak directly, bridging large-language-model reasoning with professional creative tools.
\section{Design}

\begin{figure}
        \centering
        \includegraphics[width=1\linewidth]{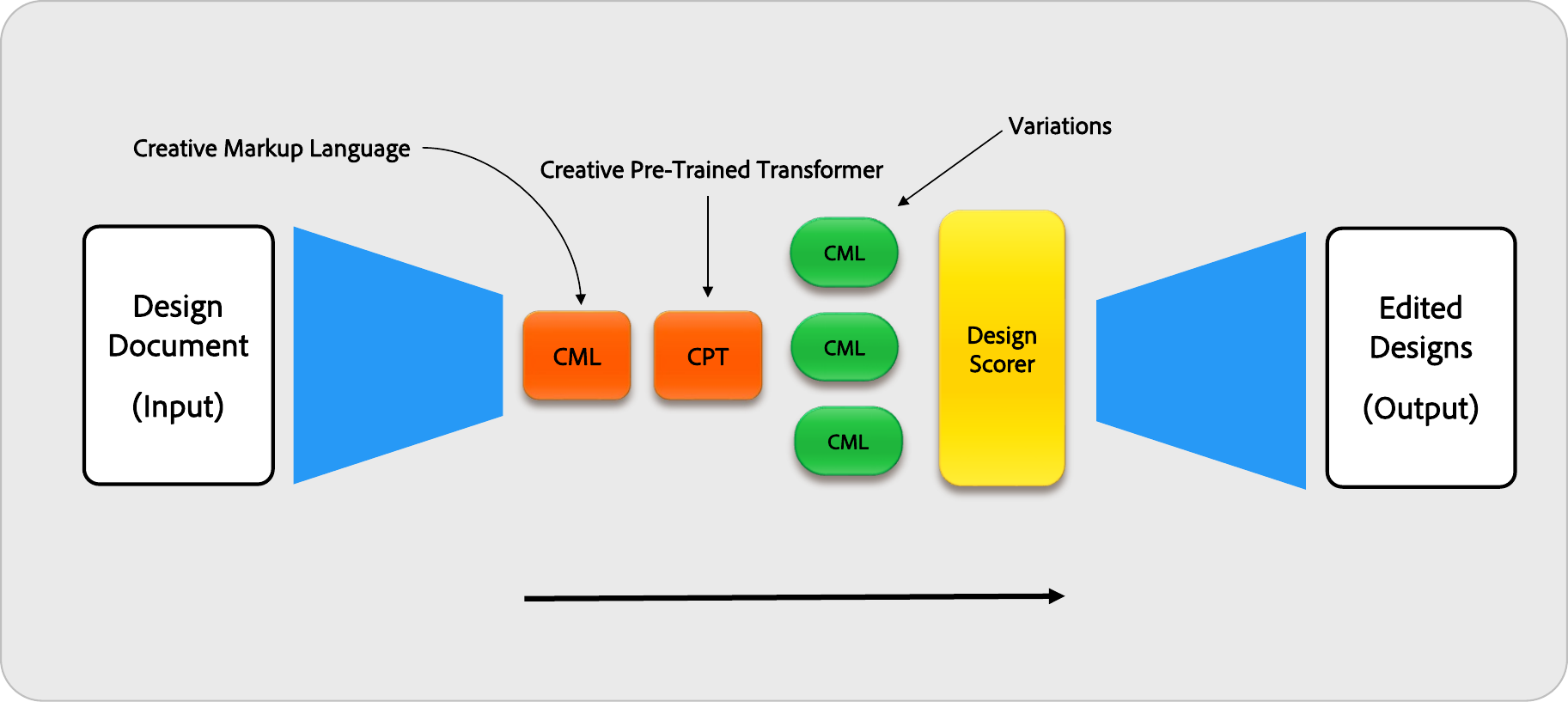}
        \caption{High-Level Overview of the Design Variations Pipeline}
        \label{fig:pipeline}
\end{figure}

Our system consists of three main components: (1) a design representation (CML) that converts graphical templates into a linear token sequence suitable for a language model, and (2) the Creative Pre-trained Transformer (CPT) model, a decoder-only transformer that is fine-tuned on the CML sequences to output design variations, and (3) a design scorer to filter designs that do not meet aesthetic standards. We also outline our training procedure for CPT, which involves masking certain attributes in the CML and letting the model predict them, and describe how we generate new variations at inference time. An overview of the pipeline is illustrated in Figure \ref{fig:pipeline}.

\subsection{Creative Markup Language (CML)}
CML (Creative Markup Language) is a domain-specific, compact, and semantically rich representation for editable design templates. It linearizes design documents into sequences of tokens while preserving their hierarchical structure. A CML document begins with global definitions (e.g., canvas size, background), followed by structured element blocks such as \texttt{<text>}, \texttt{<image>}, or \texttt{<shape>}. Each block encodes both \textbf{content} (e.g., text string, image reference) and \textbf{style/layout attributes} (e.g., font, size, color, bounds, z-index). By masking attributes such as color, font, or bounds, CML provides controllable slots for CPT to predict stylistically coherent alternatives.



Compared to raw JSON or SVG, CML offers several advantages:  
\begin{itemize}[leftmargin=*]
    \item \textbf{Machine-learning--friendly:} standardized tokens and normalized values (e.g., hex colors, canonical font names) allow efficient training.  
    \item \textbf{Context-aware:} the order and grouping of tokens preserve structural relationships, enabling models to infer design conventions (e.g., titles and subtitles sharing a style).  
    \item \textbf{Editable:} round-trips seamlessly back into production design tools, ensuring generated outputs remain fully manipulable.  
    \item \textbf{High-fidelity:} provides a near-exact reverse mapping to renderable documents, preserving layout, styling, and content integrity.  
\end{itemize}

Crucially, CML supports attribute masking (e.g., replacing color codes or font families with \texttt{<MASK\_COLOR>} or \texttt{<MASK\_FONT>}) to guide CPT in predicting stylistically coherent variations.  

\begin{verbatim}
<text id="3">
  <bounds top="239" left="210" width="600" height="385"/>
  <style alignment="center"/>
  <p>
    <content>Pizza</content>
    <style color="<MASK_COLOR_0>" font="<MASK_FONT_0>" size="<MASK_FONTSIZE_0>">
  </p>
</text>
\end{verbatim}

This masking strategy ensures predictions remain contextualized and coherent, while making CML a natural substrate for causal LLM training.

\subsection{Creative Pre-trained Transformer (CPT)}

CPT is a transformer-based language model that we fine-tuned for design generation. We start from a pre-trained decoder-only LLM with 7B parameters (Mistral-7B \citep{jiang2023mistral7b}), chosen for its strong contextual reasoning and ability to generate well-structured text such as code or JSON \citep{wei2022chain}. Since CML is designed as a flat, non-nested XML representation, it avoids the hierarchical complexity that commonly causes parsing errors in LLM-generated JSON or SVG. Each element block (e.g., <text>, <image>, <shape>) maps directly to a design entity, keeping syntax shallow yet expressive. Fine-tuning CPT on over 220K professionally authored CML templates enables it to learn this grammar and structure, producing syntactically valid XML in over 99\% of validation cases verified by schema checks. This flat schema and large-scale finetuning together yield reliably well-formed and semantically coherent design documents.

\textbf{Fine-tuning Objective.} CPT is fine-tuned on 220K professionally designed templates converted to CML using LoRA \citep{hu2021lora} for parameter-efficient adaptation. The task is formulated as masked sequence prediction: selected attributes are replaced with placeholder tokens, and the model autoregressively predicts only the masked values. These predictions are then \emph{infilled} back into the original CML at the correct positions, preserving the rest of the canvas exactly. This masking-based infilling resembles the “Fill-in-the-Middle” (FIM) paradigm \citep{bavarian2022efficient}, but differs in that our approach uses \emph{pre-defined, task-specific masks} (e.g., colors, fonts, layout) within structured XML (CML). The model outputs only the masked tokens, ensuring contextualized yet controlled design variations rather than free-form text completion. To explore different levels of consistency and diversity, we trained three variants of CPT:  
\begin{itemize}[leftmargin=*]
    \item \textbf{CPT No Association:} every masked token is independent, giving the model maximal freedom but often yielding inconsistent styles.  
    \item \textbf{CPT Local Association:} masks within the same element (e.g., all attributes in a \texttt{<text>} block such as color, font, size) share a mask ID, enforcing local consistency.  
    \item \textbf{CPT Global Association:} attributes that match across elements share the same mask ID, enforcing global coherence (e.g., multiple boxes sharing the same color). This reduces diversity slightly but improves alignment, contrast, and design consistency—ensuring that identical attributes in the original remain identical in all variations. Such consistency is especially valuable in structured visuals like infographics or figures with legends, where these associations carry semantic meaning.
(see Table~\ref{tab:model_comparison}).  
\end{itemize}

This approach is akin to sequence “inpainting”: the model sees a mostly complete XML canvas, outputs only the missing attributes, and these are seamlessly infilled into the masked positions, yielding a fully valid and renderable design document. Temperature further provides a controllable knob on creativity: lower values bias the model toward safer, brand-consistent predictions, while higher values encourage more diverse and exploratory style variations.

For illustration, consider the earlier \texttt{<text>} element. After masking font and color attributes, the model produces predictions that are then infilled back into the CML:

\begin{verbatim}
<text id="3">
  <bounds top="239" left="210" width="600" height="385"/>
  <style alignment="center"/>
  <p>
    <content>Pizza</content>
    <style color="<MASK_COLOR_0>" font="<MASK_FONT_0>" size="MASK_FONTSIZE_0"/>
  </p>
</text>

### OUTPUT:
MASK_COLOR_0: #fefcf0
MASK_FONT_0: MeowScript-Regular
MASK_FONTSIZE_0: 279
\end{verbatim}

These predicted values are inserted back into the masked positions, yielding a fully valid and renderable CML document. For brevity, only a fragment of the full CML is shown here; a complete masked-to-infilled example is provided in the Appendix.

All models were fine-tuned for 3 epochs on hundreds of thousands of professionally designed templates from an online graphic design platform, covering diverse use cases (social media posts, flyers, ads, invitations, etc.). Each template was converted to CML, and multiple masked variants were generated to expose the model to a wide range of scenarios (colors only, fonts only, or combinations including layout attributes).  

Training used a standard autoregressive loss with AdamW optimizer \citep{loshchilov2017adamw} and GELU activations \citep{hendrycks2016gelu}. Mask placeholders (e.g., \texttt{<MASK\_COLOR>}) remain in the input, and CPT learns to output only the missing values, which are then infilled into the original CML sequence. This allows the model to generate context-aware, stylistically coherent predictions while remaining faithful to the XML structure of CML.

\subsection{Variation Generation and Design Scorer}

At inference time, variations are generated by masking selected attributes in the input CML and letting CPT predict replacements, which are then infilled to produce a complete, editable design. Users can mask \emph{color, font, layout, or any combination of these}, giving direct control over which aspects of the design are modified:

\begin{itemize}[leftmargin=*]
    \item \textbf{Color variations:} color attributes  (background color, text color, effect color, etc.) are masked and repainted by CPT, typically yielding coherent palettes rather than arbitrary colors.  
    \item \textbf{Font variations:} font attributes (family, size, leading, tracking, etc.) are masked, allowing CPT to propose new typographic styles while preserving layout and content.
    \item \textbf{Layout variations:} positional attributes (e.g., coordinates, sizes) can be masked to explore alternative arrangements. This is a more challenging problem, but CPT can still maintain balance in simple cases, suggesting it has learned basic spatial relationships \citep{lee2020neural}.
   \item \textbf{Effect, brand-aware, and other variations:} the same masking strategy naturally extends beyond color, font, and layout. For instance, effect attributes (e.g., duotone, colorize, tint) can be masked to explore stylistic treatments, brand-aware variations can be guided by constraints in a dedicated \texttt{<brand>} section of the CML, and additional attributes can be incorporated as needed—making this a general, extensible framework for creative control.


See Appendix~\ref{color_font_ex}\ref{layout_ex} \ref{brand-ex} for qualitative illustrations of variation types.

\end{itemize}
The CML → design conversion is deterministic and implemented via a diff-and-apply pipeline. We first compute a structured diff between the original and infilled CML, translate these changes into atomic document-edit operations (e.g., style, font, or layout updates) bound to stable element IDs, and apply them through the platform’s rendering APIs to produce updated renditions. Because each edit is schema-validated and uniquely targeted, the result is visually and structurally identical to the baseline and remains fully editable. When no base document exists, we reconstruct the design deterministically by reversing the CML mapping to restore elements, geometry, styles, and asset references—ensuring a consistent baseline for subsequent diff-and-apply updates.

To guarantee quality, we employ a Design Scorer that leverages GPT-4o \citep{openai2024gpt4o} to filter and rank rendered variations, assessing both usability and aesthetic soundness.

\subsubsection{Filtering and Failure Detection}
Despite strong performance, CPT can still produce occasional failures due to the inherent difficulty of modeling aesthetics, the limitations of text-only representations, and imperfections in data distribution. To mitigate these issues, we designed a system that automatically detects common failure modes before presenting the generated design variations to end users.

The filtering component operates on two levels: CML-based and rendition-based. This two-stage process enables early rejection of low-quality generations before costly rendering, while still ensuring rigorous post-render checks. The component is also customized to detect failures according to different generation modes. For conciseness, the following discussion focuses on the generation mode for color and font variations.

\textbf{CML-based filtering.} In the case of color and font variations, generations that lack sufficient diversity—such as those with colors or fonts too similar to either the original design or to other generated variations—are discarded.

\textbf{Rendition-based filtering.} This stage operates on rendered images of both the original template and the generated variations. Leveraging GPT-4o’s multimodal capabilities, the system evaluates design quality using two targeted metrics specific to color and font variations.

\begin{itemize}[leftmargin=*]
    \item \textbf{Color Contrast Filtering:} Both the original and modified design images are input to GPT-4o with a specialized color contrast prompt. The model evaluates whether the color variations have introduced readability issues such as unreadable, faded, or missing text and design elements due to poor contrast. Each variation is categorized as either \textit{pass} (no visibility issues detected) or \textit{fail} (important elements are missing or unreadable). Only variations receiving a \textit{pass} classification are retained for further consideration.
     \item \textbf{Alignment Filtering:} Using the original and modified images, GPT-4o checks for alignment issues introduced by the variation process with a dedicated alignment prompt. Importantly, the model flags only newly introduced alignment problems, not pre-existing flaws in the original template. Results are classified as \textit{pass} (no usability or layout issues) or \textit{fail} (obvious misalignment affecting usability or aesthetics). Variations that fail this stage are removed from the candidate set.
\end{itemize}

\subsubsection{Aesthetic Ranking and Diversity Maximization}

After filtering, the remaining variations for each template undergo a ranking process designed to balance aesthetics and diversity. Using GPT-4o with a diversity-emphasized prompt, variations are ranked to ensure the final set collectively presents a broad range of color palettes and stylistic approaches while maintaining quality standards.

This scorer ensures both quality assurance and diversity optimization: users receive variations that are technically sound while also offering stylistically meaningful alternatives. These results demonstrate that in-context learning with GPT-4 can reliably function as a design scorer \citep{haraguchi2024can}.

\section{Results}

\subsection{Evaluation Methodology}

To assess the quality of CPT’s generated variations, we use a \emph{three-pronged evaluation framework} that balances automated analysis with human-centered judgments: \textbf{Automated Heuristic Metrics}—quantitative checks used for checkpoint selection; \textbf{Human Evaluation}—thumbs-up/down ratings with comments; and \textbf{Qualitative Golden-Set Analysis}—manual inspection of challenging templates.

\subsubsection{Automated Heuristic Metrics and Results}
We designed a suite of heuristic-based metrics that capture critical design principles such as spatial layout integrity, text readability, and visual contrast. While useful for guiding model selection, these metrics cannot fully capture aesthetic quality. All metrics are reported as \textit{chosen rates} (percentage of designs meeting a minimum quality threshold), following \citep{inoue2024opencole, horita2024retrieval}.

\begin{itemize}[leftmargin=*, itemsep=2pt, topsep=0pt]
    \item \textbf{General Overlap}: Measures the percentage of designs where elements do not inappropriately overlap, preserving visual clarity and structure.
    \item \textbf{Text Overflow}: Evaluates whether text fits within its bounding box without clipping horizontally or vertically.
    \item \textbf{Text Over Boundary}: Assesses the percentage of text elements within canvas boundaries.
    \item \textbf{Text Line Overlap}: Measures whether text lines interfere with other elements.
    \item \textbf{Color Contrast}: Evaluates if text has sufficient contrast against the background for readability.
    \item \textbf{Overall Chosen Rate}: A composite metric that requires passing all quality checks, serving as a conservative indicator of immediately usable designs.
\end{itemize}

\noindent Table~\ref{tab:model_comparison} presents quantitative results of this evaluation. All reported results correspond to the CPT model trained for 3 epochs and evaluated at temperature 0.8, selected as the checkpoint with the lowest validation loss and strongest performance on heuristic metrics. 

\begin{table}[h]
\centering
\caption{CPT model performance on heuristic metrics (Chosen Rate \%). Best non-human values in \textbf{bold}.}
\label{tab:model_comparison}
\resizebox{\linewidth}{!}{%
\begin{tabular}{lcccccc}
\toprule
\textbf{Model} & \makecell{\textbf{Overall}\\\textbf{Chosen Rate}} &
\makecell{\textbf{General}\\\textbf{Overlap}} &
\makecell{\textbf{Text}\\\textbf{Overflow}} &
\makecell{\textbf{Text}\\\textbf{Over Boundary}} &
\makecell{\textbf{Text}\\\textbf{Line Overlap}} &
\makecell{\textbf{Color}\\\textbf{Contrast}} \\
\midrule
Human Templates (Gold Standard) & 80.7 & 97.8 & 93.5 & 99.7 & 98.8 & 94.6 \\
\midrule
CPT Global Association & \textbf{58.3} & \textbf{93.5} & 63.7 & \textbf{99.6} & \textbf{98.1} & \textbf{41.3} \\
CPT Local Association  & 50.9 & 93.3 & 61.3 & \textbf{99.6} & 98.0 & 35.4 \\
CPT No Association     & 35.0 & 92.5 & \textbf{66.2} & 99.3 & 96.5 & 26.3 \\
\bottomrule
\end{tabular}%
}
\end{table}

\subsubsection{Human Evaluation}

To complement the automated metrics, we conducted a human evaluation with professional designers and engineers. In total, we collected \textbf{2974} ratings on variations filtered and ranked by the Design Scorer: \textbf{2698} thumbs up (\textbf{90.7\%}) and \textbf{276} thumbs down (\textbf{9.3\%}).

\begin{figure}[h!]
    \centering
    \includegraphics[height=0.25\textheight, keepaspectratio]{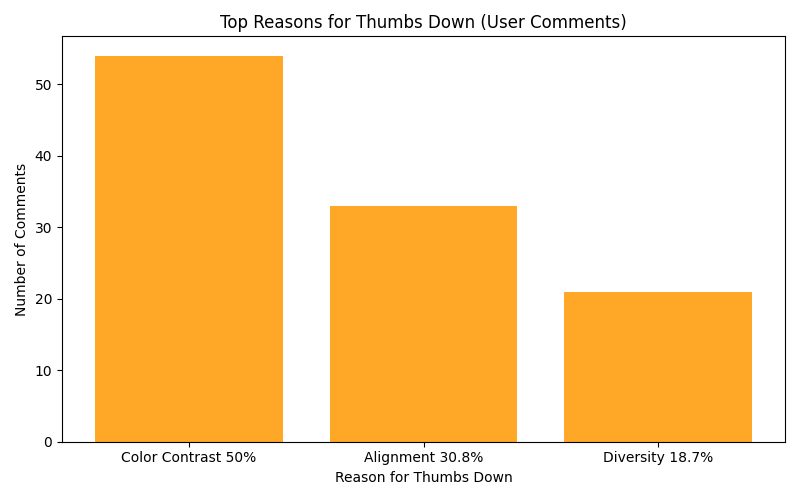}\par\vspace{1mm}
    \caption{Human evaluation results}
    \label{fig:cpt_human_eval}
\end{figure}

We also collected \textbf{107} free-form comments explaining thumbs-down ratings. Figure~\ref{fig:cpt_human_eval} summarizes the distribution of these comments, with the most frequent concerns being insufficient color contrast \textbf{(50\%)}, misalignment issues \textbf{(30.8\%)}, and lack of stylistic diversity \textbf{(18.7\%)}.

\paragraph{Validation of the Design Scorer.}
We validated the Design Scorer, which contains GPT-based filtering on color contrast and alignment, by corroborating its pass/fail judgments on those by human ratings (see Table \ref{tab:cm_best}); detailed prompt comparisons are in Appendix ~\ref{prompt}

\begin{table}[h]
\centering
\caption{Confusion matrix comparing GPT judgments (Pass vs. Fail; with Borderline treated as Pass) to human thumbs (Up vs. Down) for the best-performing prompt. Cells show \emph{count} (row\%).}
\label{tab:cm_best}
\begin{tabular}{@{}lcc@{}}
\toprule
 & \textbf{GPT: Fail} & \textbf{GPT: Pass} \\
\midrule
\textbf{Human: Down} & \makecell{561 \\ (18.9\%)} & \makecell{665 \\ (22.4\%)} \\
\textbf{Human: Up}   & \makecell{201 \\ (6.8\%)} & \makecell{1547 \\ (52.0\%)} \\
\bottomrule
\end{tabular}
\end{table}

\subsubsection{Qualitative Evaluation on a Golden Set}

Finally, we conduct qualitative comparisons using a curated “golden set” of challenging designs. These include edge cases such as dense layouts, extreme aspect ratios, and complex typography. We also include cases requiring world knowledge and design conventions, such as seasonal color themes (e.g., orange–purple for Halloween, green–red for Christmas) and domain-specific typography choices (e.g., playful fonts for children’s events, serif fonts for formal announcements). 

For each case, we manually compare CPT’s outputs against human-designed references, enabling detailed analysis of strengths, weaknesses, and failure modes that aggregate scores cannot capture. 



\subsection{Analysis and Key Findings}


\begin{itemize}[leftmargin=*]
    \item \textbf{Human upper bound:} Human templates achieve 80.7\% chosen rate, setting the maximum benchmark for this evaluation.
    \item \textbf{Association matters:} Association-based models strongly outperform the no-association baseline (58.3\% vs.\ 35.0\%).
    \item \textbf{Color contrast:} Remains the largest gap (41.3\% vs.\ 94.6\%), though GPT-based filtering boosts satisfaction to 90.7\%.
\end{itemize}



These results show that while CPT faces challenges in color contrast and certain aspects of typography, it reliably preserves structure and consistency. Automated metrics alone can underestimate aesthetic quality, but when combined with human evaluation and qualitative golden-set inspection, they provide a more complete picture of CPT’s strengths and gaps. The Design Scorer further ensures that surfaced variations are stylistically diverse, structurally sound, and well-received—competitive with recent multimodal design generation systems \citep{cheng2025graphic, lin2025elements}.

\section{Conclusions}



We presented CPT, a novel approach for generating editable design variations using fine-tuned language models and a structured design representation. Our contributions include the Creative Markup Language (CML), which enables compact and semantically rich encoding of design templates, and the Creative Pre-trained Transformer (CPT), which predicts contextually appropriate style attributes through masked sequence prediction. In addition, we introduce a filtering and ranking pipeline powered by GPT-4o that detects contrast and alignment issues, ranks variations for aesthetics and diversity, and ensures only high-quality outputs reach users. 

We also design a comprehensive evaluation framework that combines automated heuristics, GPT-based assessments, and human judgments, offering a reliable picture of both functional and aesthetic quality. By producing fully editable outputs rather than static images, CPT addresses a key limitation of existing generative design tools. Fine-tuned on professionally designed templates, the system generates coherent color and font variations while maintaining design consistency, demonstrating how LLM-driven approaches can balance creativity with reliability in design generation.

Our work opens new possibilities for AI-assisted design tools that preserve creative control while accelerating content production, potentially transforming how designers approach template variation and personalization at scale.




\subsection{Future Work}
We see three directions. (1) \emph{Multimodal + scale:} integrate visual signals from embedded photos and design assets and train at larger scale, so CPT can reason about text–background interactions and layering, mitigating color-contrast and alignment issues inherent to text-only CML. (2) \emph{Richer CML + controllable inference:} extend CML with constraints and design relationships (e.g., component groupings) to enforce consistency, and expose these constraints at inference—incorporating user preferences and brand guidelines for personalized, brand-consistent outputs. (3) \emph{Better evaluation + training loops:} develop metrics and stronger template-to-CML conversion to detect failures (color contrast, alignment, overlap) more accurately in the Design Scorer and during training, enabling reinforcement-style training to further improve quality.

\section*{Acknowledgements}
We would like to thank the members of Adobe Enterprise Search engineering team including Rahul Gandhi, Vishwesh Nayak, and Nandaja Ananthanarayanan for building data and rendering pipelines. We also would like thank Tracy King, Gaurav Kukal and Vipul Dalal for their support and encouragement of this project.

{\small
\bibliographystyle{abbrvnat}
\bibliography{egbib}
}



\clearpage
\appendix
\section{Appendix}


\subsection{GPT Evaluation Prompts and Results} \label{prompt}

To assess reliability of the GPT-based filtering pipeline, we iterated over multiple prompt formulations. Below, we report detailed results for \texttt{prompt\_v1} and the improved \texttt{prompt\_v2}, evaluated against human thumbs up/down judgments for color contrast and alignment combined. Borderline cases are considered as passes. 

\begin{table}[h]
\centering
\caption{Confusion matrices comparing GPT judgments (Pass vs. Fail; Borderline = Pass) to human thumbs for \texttt{prompt\_v1} and \texttt{prompt\_v2}. Cells show \emph{count} (row\%).}
\label{tab:cm_v1v2}
\begin{minipage}{0.48\linewidth}
\centering
\textbf{Prompt\_v1} \\
\vspace{2mm}
\begin{tabular}{@{}lcc@{}}
\toprule
 & \textbf{GPT: Fail} & \textbf{GPT: Pass} \\
\midrule
\textbf{Human: Down} & \makecell{686 \\ (23.1\%)} & \makecell{1350 \\ (45.4\%)} \\
\textbf{Human: Up}   & \makecell{76 \\ (2.5\%)}   & \makecell{862 \\ (29.0\%)} \\
\bottomrule
\end{tabular}
\end{minipage}
\hfill
\begin{minipage}{0.48\linewidth}
\centering
\textbf{Prompt\_v2} \\
\vspace{2mm}
\begin{tabular}{@{}lcc@{}}
\toprule
 & \textbf{GPT: Fail} & \textbf{GPT: Pass} \\
\midrule
\textbf{Human: Down} & \makecell{561 \\ (18.9\%)} & \makecell{665 \\ (22.4\%)} \\
\textbf{Human: Up}   & \makecell{201 \\ (6.8\%)}  & \makecell{1547 \\ (52.0\%)} \\
\bottomrule
\end{tabular}
\end{minipage}
\end{table}

\subsection*{Prompt\_v1} 
\textbf{Color contrast:} ``You will be provided with an original template (first image) and a modified template (second image) with changes in color for certain components. Determine if the second template has any missing or unreadable text/elements due to low contrast. Pay attention to all texts and object regardless of their size, maintaining a very high threshold for visibility. What you can do it try to recognize all texts in the first image, then do the same for the second image, independently. Compare if all extracted texts match one another. If any text/object is missing in the second image, response 'yes'. If ALL the design elements are readable, flag them as 'no.' If there are issues, specify which parts of the design lack sufficient contrast.''  

\textbf{Alignment:} ``You will be provided with an original template (first image) and a modified template (second image) with changes in color and fonts. Determine if the second template has any text placement different from the original, resulting in misalignment with other objects and reduced layout harmony. Examine closely at each text, pay attention to any overlapping of texts and objects, off-center placement of texts, text extending beyond canvas or containers, text too small and not taking enough space in its designated containers, broken alignment with respect to other texts in the design, reduced readability due to text misplacement. Only respond 'yes' if any behavior listed above is found. If none of the above issues are found flag them as 'no.' If there are issues, specify which parts of the design has text misalignment issues.''  

\textbf{Color contrast.} ``You will be provided with two images: the original\_design\_template and a modified\_design\_template with changes in color, font \& font size. Your task is to determine whether the modified\_design\_template has any missing, unreadable, or significantly faded text or design elements due to very poor color contrast. Focus on real-world readability: flag only if the contrast in the modified\_design\_template makes the text significantly harder to perceive compared to the original. Minor reductions in clarity that do not affect readability should not be flagged. The original\_design\_template is provided for context to understand existing visibility levels. We are interested in knowing if the modified\_design\_template introduced new visibility issues that could impact user experience. Ignore spelling corrections or differences --- evaluate visibility only. Even small text or visually distinct elements such as headers, badges, banners, or call-to-action buttons must be clearly readable against their background. Do not assume visibility based on typical layout or expected content. Text that remains clearly visible---even with bold, unconventional, or stylistic color choices---should be considered acceptable. Categorize your response using one of the following three buckets: 1. `clear\_pass' --- all elements are present and fully readable with no visibility concerns; 2. `borderline' --- some text or elements are slightly harder to perceive but are readable and might still be acceptable; 3. `clear\_fail' --- important text or elements are missing, unreadable, or significantly harder to perceive. Return your answer using this format: `<bucket>: <short explanation>'. Example1: clear\_fail: The red text in the footer is unreadable against the dark background. Example2: borderline: text abc is slightly harder to read against the light background, but it might still be acceptable. Example3: clear\_pass: All elements are clearly visible and readable. \textbf{Do not evaluate for alignment or minor layout adjustments or pre-existing flaws in the original design.}''

\textbf{Alignment.} ``You will be provided with two images: the original\_design\_template and a modified\_design\_template with changes in color, font, font size or minor layout adjustments. Your task is to determine whether the modified\_design\_template introduces \textbf{any new alignment issues} that negatively impact layout quality, visual balance or user perception. \textbf{Ignore the color contrast issues and spelling mistakes}. \textbf{Focus strictly on new alignment problems introduced in the modified\_design\_template which were not present in the original\_design\_template}, such as: Text or elements overlapping with each other; Text misaligned enough to appear visually far disconnected from related elements; Text or objects extending outside canvas area; Words split awkwardly across lines (e.g., a single word broken between lines). Categorize your response using \textbf{one of these buckets}: 1. `clear\_pass' --- Layout remains visually coherent with no usability concerns 2. `borderline' --- Some misalignments are present but may still be acceptable 3. `clear\_fail' --- Clear misalignment issues that disrupt readability or balance. \textbf{Respond using this exact format}: `<bucket>: <short explanation>'. \textbf{Examples:} Example 1: clear\_fail: Footer text is misaligned and overlaps with page number. Example 2: borderline: Sidebar content is slightly shifted but still understandable. Example 3: clear\_pass: All design elements are properly aligned and balanced. \textbf{Do not evaluate color, text contrast, spelling, or pre-existing flaws in the original design.}''

\paragraph{Analysis}  
Compared to \texttt{prompt\_v1}, \texttt{prompt\_v2} introduced:  
\begin{itemize}[leftmargin=*]
    \item Clearer instructions to \emph{ignore non-relevant issues} (e.g., spelling, color when evaluating alignment).  
    \item A structured triage scheme (\texttt{clear\_pass}, \texttt{borderline}, \texttt{clear\_fail}) instead of binary labels.  
    \item Emphasis on real-world readability and user perception, rather than absolute pixel-level differences.  
\end{itemize}

This change reduced false positives substantially (from 45.4\% to 22.4\%) and improved true positives (from 29.0\% to 52.0\%), at the cost of slightly higher false negatives. The result is a more reliable and user-aligned filter, ensuring that genuinely usable designs are surfaced while only discarding those with meaningful quality issues.

\subsection{Qualitative Examples of Color and Font Variations} \label{color_font_ex}
\begin{figure}[ht]
    \centering
    \includegraphics[width=\linewidth]{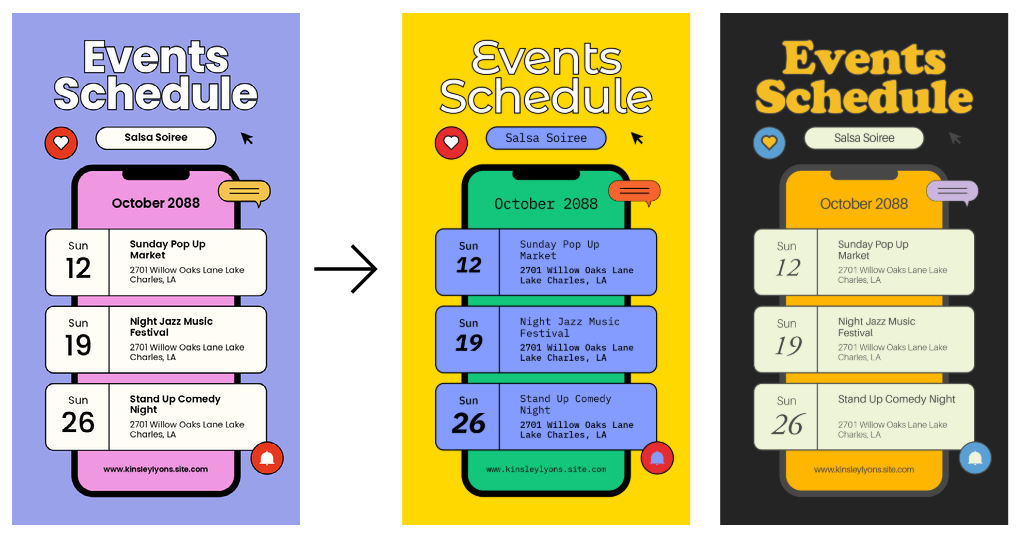}\par\vspace{2mm}
    \includegraphics[width=\linewidth]{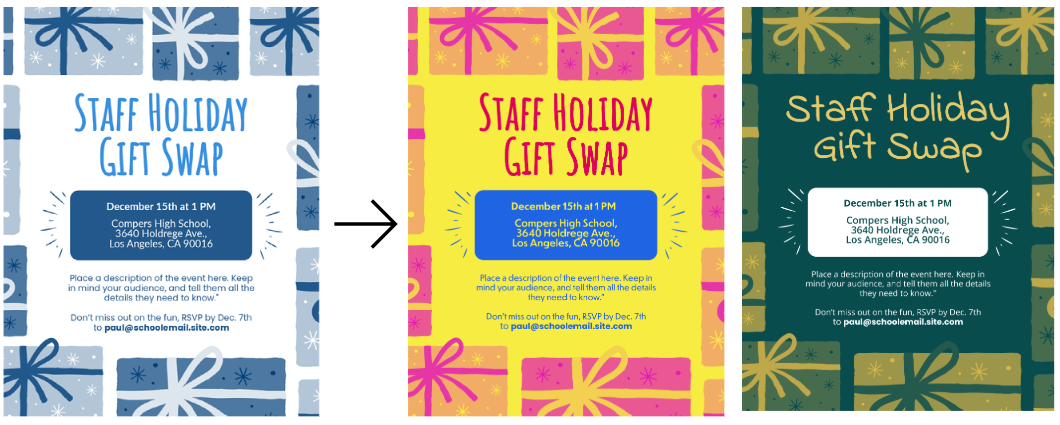}\par\vspace{2mm}
    \includegraphics[width=\linewidth]{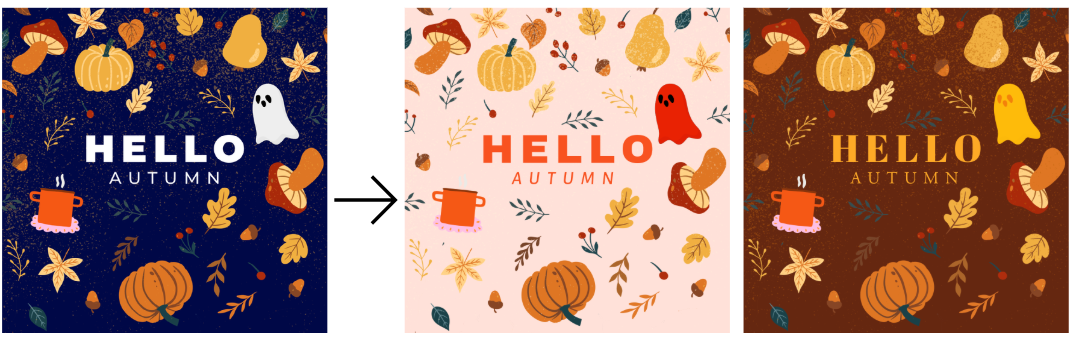}\par\vspace{2mm}
    \includegraphics[width=\linewidth]{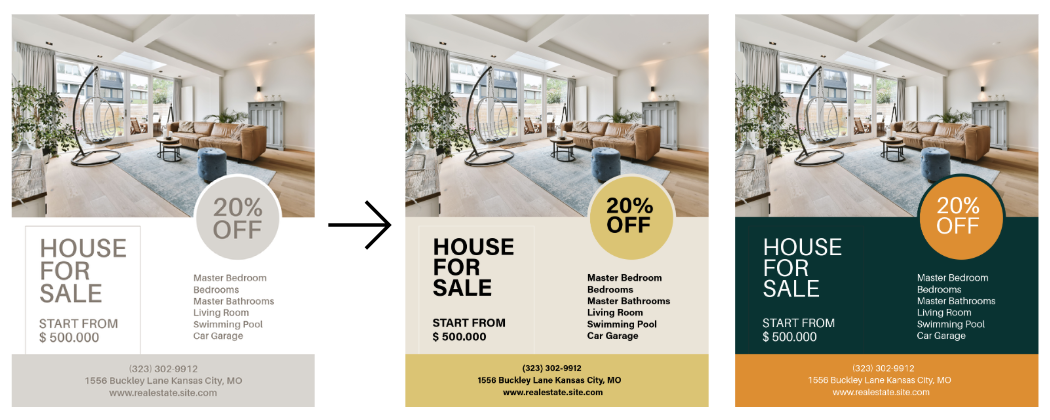}\par\vspace{2mm}
    \caption{Our CPT model generates stylistic variations (right) from an original design (left). Each row shows a different example with either font or color variation. Notably, Example 2 illustrates the use of world knowledge to select a Halloween-inspired color palette, while Example 3 demonstrates context-aware typography, where playful fonts are chosen to match the event theme.
}
    \label{fig:cpt_examples}
\end{figure}
\FloatBarrier

\subsection{Examples for Layout Variations}\label{layout_ex}
\begin{figure}[ht]
    \centering
    \includegraphics[height=0.22\textheight, keepaspectratio]{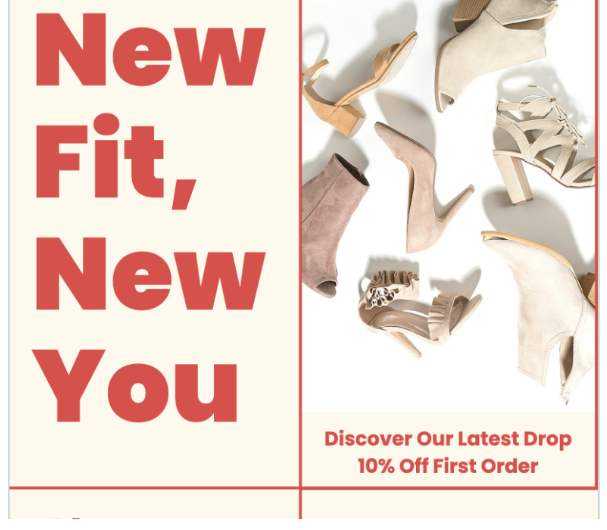}\par\vspace{2mm}
    \includegraphics[height=0.25\textheight]{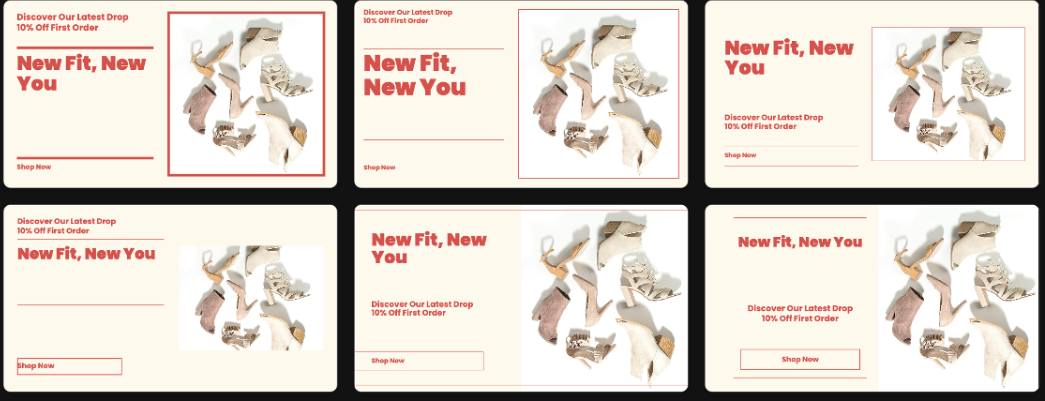}\par\vspace{2mm}
    \includegraphics[height=0.33\textheight]{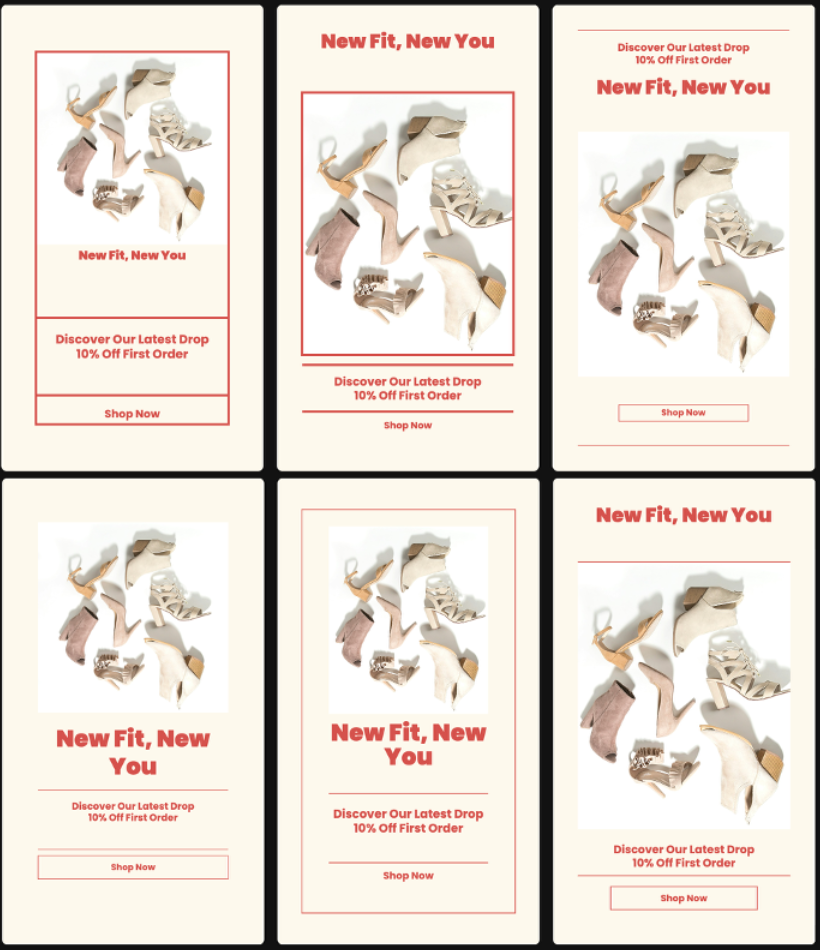}\par\vspace{2mm}
    \caption{Examples of layout variations generated from a single template: the original square format (1:1) in the first row, adapted to a YouTube thumbnail in the second row, and to an Instagram Story in the third row.}

    \label{fig:cpt_layout_examples}
\end{figure}

\subsection{Examples for Brand-Aware Color-Font Variations} \label{brand-ex}
\begin{figure}[ht]
    \centering
    \includegraphics[height=0.45\textheight]{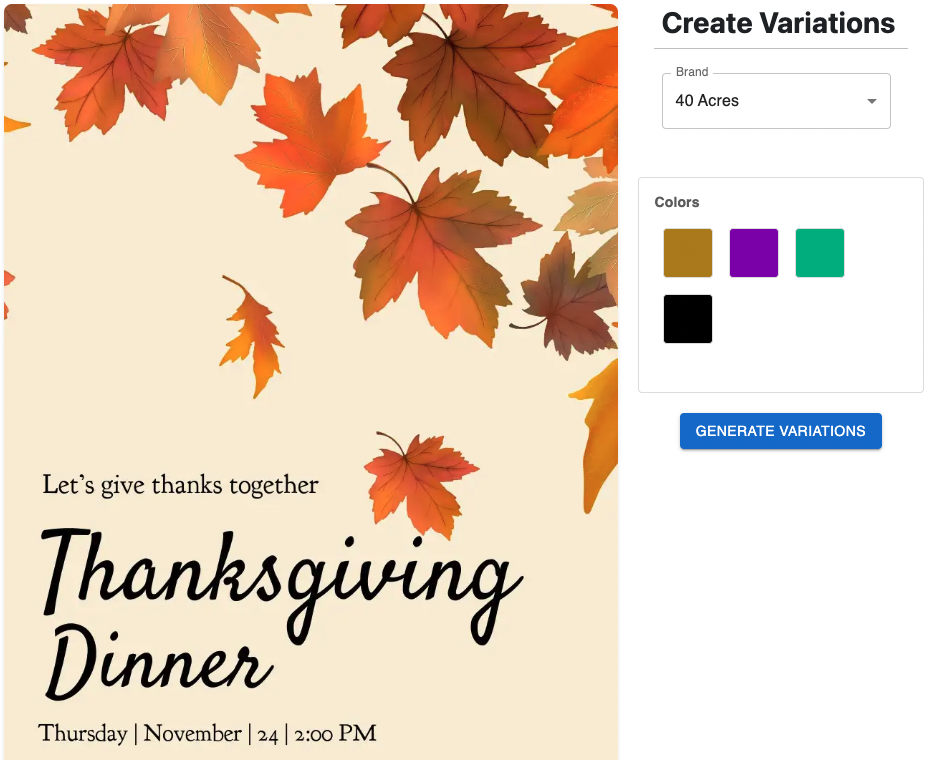}\par\vspace{2mm}
    \includegraphics[height=0.45\textheight]{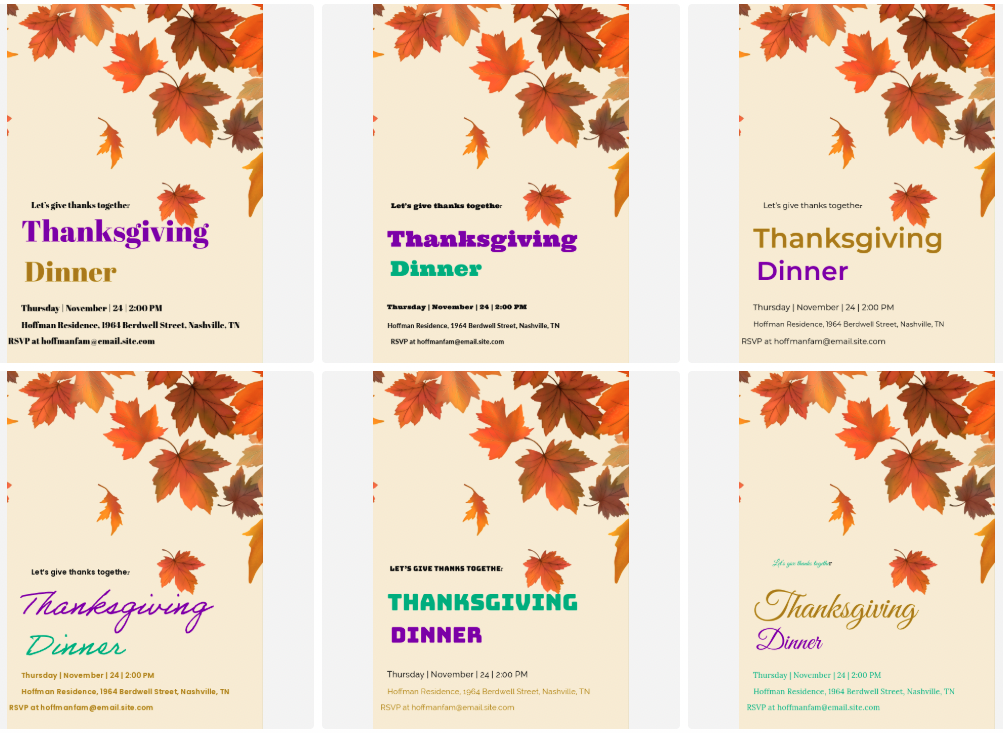}\par\vspace{2mm}
    \caption{CPT applies a brand preset (colors + fonts) to generate variations that stay editable, contextually coherent, and consistent with brand identity.}

    \label{fig:cpt_brand_aware_examples}
\end{figure}

\FloatBarrier
\subsection{Full Example of CML Representation}

Below we include a complete Creative Markup Language (CML) specification for one of our generated designs. This highlights the structured representation of colors, fonts, shapes, and layout.

\begin{lstlisting}[language=XML, basicstyle=\ttfamily\scriptsize, breaklines=true]
<cml v="3" numberPages="1">
    <page width="2550" height="3300" format="flyer" layoutID="unknown">
        <brand>
            <color value="#eceae3" />
            <color value="#400e04" />
            <color value="#a3665c" />
            <color value="#bd521d" />
            <color value="#400e05" />
            <color value="#e0e0e0" />
            <font name="Rubik-SemiBold" />
            <font name="Poppins-Regular" />
            <font name="Satisfy-Regular" />
            <font name="Poppins-Black" />
            <font name="Rubik-Italic" />
        </brand>
        <background type="color" id="0" entityId="b04d9fa0-6c15-459d-bb5f-fd4199c18616">
            <bounds top="0" left="0" width="2550" height="3300" rotation="0" z-index="0" />
            <style color="#fff6eb" />
        </background>
        <text id="1" entityId="26384d45-34d4-419f-9cbc-285588da6618">
            <bounds top="246" left="426" width="371" height="79" rotation="0" z-index="3" />
            <style alignment="left" layout="dynamic" />
            <p>
                <content>
                    E m e n
                </content>
                <style leading="1.2" color="#008045" font="Novecentosansnarrow-Bold" size="116" tracking="0" opacity="1" underline="false" fontSize="48" />
            </p>
        </text>
        <text id="2" entityId="0ed7c280-99ed-41d1-9b02-b77602ec6683">
            <bounds top="588" left="255" width="1287" height="58" rotation="0" z-index="4" />
            <style alignment="left" layout="dynamic" />
            <p>
                <content>
                    Please join us for our annual
                </content>
                <style leading="1.2" color="#a8493f" font="Muli-Regular" size="100" tracking="0" opacity="1" underline="false" fontSize="90" />
            </p>
        </text>
        <text id="3" entityId="2e175ef3-8f2e-465e-abbe-9eac674c1707">
            <bounds top="733" left="200" width="1850" height="498" rotation="0" z-index="5" />
            <style alignment="left" layout="autoWidth" />
            <p>
                <content>
                    Thanksgiving
                </content>
                <style leading="1.2" color="#782010" font="Allura-Regular" size="374" tracking="0" opacity="1" underline="false" fontSize="415" />
            </p>
        </text>
        <text id="4" entityId="23885bd6-6c78-4ee7-913e-6c3f7bfde03c">
            <bounds top="1650" left="227" width="1541" height="232" rotation="0" z-index="6" />
            <style alignment="left" layout="autoHeight" />
            <p>
                <content>
                    Friday, Nov. 17th    |    12-2pm\r
                </content>
                <style leading="1.69" color="#8a362c" font="Muli-Bold" size="84" tracking="0" opacity="1" underline="false" fontSize="86" />
            </p>
            <p>
                <content>
                    Conference Room A
                </content>
                <style leading="1.69" color="#8a362c" font="Muli-Regular" size="84" tracking="0" opacity="1" underline="false" fontSize="86" />
            </p>
        </text>
        <text id="5" entityId="1193cfe6-366a-4a6d-808d-5e48a8fafa4e">
            <bounds top="2134" left="232" width="716" height="214" rotation="0" z-index="7" />
            <style alignment="left" layout="autoHeight" />
            <p>
                <content>
                    Don't forget to bring a dish!
                </content>
                <style leading="1.2" color="#782010" font="Muli-Regular" size="88" tracking="0" opacity="1" underline="false" fontSize="92" />
            </p>
        </text>
        <text id="6" entityId="8dbdd45b-6bdc-4e9c-a456-483b941ee8ea">
            <bounds top="338" left="422" width="375" height="54" rotation="0" z-index="10" />
            <style alignment="left" layout="autoWidth" />
            <p>
                <content>
                    Health Solutions
                </content>
                <style leading="1.2" color="#008045" font="NotoSans-Regular" size="48" tracking="0" opacity="1" underline="false" fontSize="46" />
            </p>
        </text>
        <text id="7" entityId="2498cdf5-2636-455b-837f-dded0229104e">
            <bounds top="1094" left="107" width="1465" height="498" rotation="0" z-index="11" />
            <style alignment="left" layout="autoHeight" />
            <p>
                <content>
                    Potluck
                </content>
                <style leading="1.04" color="#782010" font="Allura-Regular" size="374" tracking="0" opacity="1" underline="false" fontSize="415" />
            </p>
        </text>
        <image id="8" entityId="a4bed395-e000-4b23-a4c9-7079be7189bc" sourceType="designAsset" sourceId="529444607">
            <bounds top="1417" left="1073" width="1819" height="1801" rotation="0" z-index="13" />
            <content>
                watercolor pumpkin clipart
            </content>
            <style blendMode="normal" hasAlpha="true" />
            <colorGrid c1="#ffffff" c2="#ffffff" c3="#ffffff" c4="#ffffff" c5="#ebd286" c6="#ffffff" c7="#ffffff" c8="#ffffff" c9="#ffffff" />
            <effect name="shape" type="Rectangle" shape="" />
        </image>
        <image id="9" entityId="f49e9351-17e2-4db3-8e43-6f888f50c856" sourceType="designAsset" sourceId="546836347">
            <bounds top="-1022" left="2735" width="2304" height="1665" rotation="90" z-index="1" />
            <content>
                autumn leaves background vector | price 1 credit usd $1
            </content>
            <style blendMode="normal" hasAlpha="true" />
            <colorGrid c1="#ca4634" c2="#ffffff" c3="#ffffff" c4="#fbd278" c5="#ffffff" c6="#ffffff" c7="#c67029" c8="#ffffff" c9="#ffffff" />
            <effect name="shape" type="Rectangle" shape="" />
        </image>
        <image id="10" entityId="b18d5b08-4c33-44fe-ba86-09e71c170686" sourceType="designAsset" sourceId="546836347">
            <bounds top="2473" left="2715" width="1397" height="3408" rotation="90" z-index="12" />
            <content>
                autumn leaves background vector | price 1 credit usd $1
            </content>
            <style blendMode="normal" hasAlpha="true" />
            <colorGrid c1="#ca4634" c2="#ffffff" c3="#ffffff" c4="#fbd278" c5="#ffffff" c6="#ffffff" c7="#c67029" c8="#ffffff" c9="#ffffff" />
            <effect name="shape" type="Rectangle" shape="" />
        </image>
        <shape id="11" type="composite" entityId="b1f0a839-6464-4ef4-8536-9e1f8c677d74" sourceType="Adobe Stock" sourceId="596680639">
            <bounds top="261" left="345" width="53" height="106" rotation="53" z-index="9" />
            <content>
                a pill icon on a white background
            </content>
            <search>
                pill icon
            </search>
            <style opacity="1" color="#008045" strokeColor="#008045" strokePosition="center" strokeWidth="1" strokeDashGeometryType="solid" />
        </shape>
        <shape id="12" type="Rectangle" entityId="2b32a5cc-476d-4f1e-98c2-2b67080e6f7e">
            <bounds top="94" left="-89" width="1162" height="594" rotation="0" z-index="2" />
            <style opacity="1" color="#fff6eb" strokeColor="#e0e0e0" strokePosition="center" strokeWidth="0" strokeMiter="10" strokeDashGeometryType="solid" />
        </shape>
        <shape id="13" type="Ellipse" entityId="99c6361e-2f62-4b27-97db-5714d0276ce7">
            <geometry rx="70" ry="70" />
            <bounds top="240" left="245" width="149" height="149" rotation="0" z-index="8" />
            <style opacity="1" color="transparent" strokeColor="#008045" strokePosition="center" strokeWidth="8.14" strokeMiter="10" strokeDashGeometryType="solid" />
        </shape>
        <shape id="14" type="Line" entityId="f6c37c37-c479-4311-a29a-f8b4705f2f6c">
            <geometry startX="0" startY="0" endX="765" endY="0" />
            <bounds top="2007" left="232" width="765" height="4" rotation="0" z-index="14" />
            <style opacity="1" color="transparent" strokeColor="#782010" strokePosition="center" strokeWidth="4" strokeMiter="10" strokeDashGeometryType="solid" />
        </shape>
    </page>
</cml>
\end{lstlisting}





\end{document}